\documentclass[letterpaper, 10 pt, conference]{ieeeconf}  

\IEEEoverridecommandlockouts                              

\usepackage{graphicx} 
\usepackage{amsmath} 
\usepackage{cite}

\usepackage[symbol]{footmisc}

\usepackage{url}
\usepackage{booktabs}
\usepackage{multirow}
\usepackage{hyperref}
\hypersetup{
  colorlinks = true, 
  urlcolor   = magenta, 
  linkcolor  = blue, 
  citecolor  = red 
}

\title{\LARGE \bf
Code-as-Symbolic-Planner: Foundation Model-Based Robot Planning via Symbolic Code Generation
}

\author{Yongchao Chen$^{1,2}$, Yilun Hao$^{1}$, Yang Zhang$^{3}$, and Chuchu Fan$^{1}$
\thanks{$^{1}$Massachusetts Institute of Technology. yilunhao@mit.edu, chuchu@mit.edu }
\thanks{$^{2}$Harvard University. yongchaochen@fas.harvard.edu}
\thanks{$^{3}$MIT-IBM Watson AI Lab. yang.zhang2@ibm.com}
}

\begin{document}

\maketitle
\thispagestyle{empty}
\pagestyle{empty}

\begin{abstract}
Recent works have shown great potential of Large Language Models (LLMs) in robot task and motion planning (TAMP). Current LLM approaches generate text- or code-based reasoning chains with sub-goals and action plans. However, they do not fully leverage LLMs' symbolic computing and code generation capabilities. Many robot TAMP tasks involve complex optimization under multiple constraints, where pure textual reasoning is insufficient. While augmenting LLMs with predefined solvers and planners improves performance, it lacks generalization across tasks. Given LLMs' growing coding proficiency, we enhance their TAMP capabilities by steering them to generate code as symbolic planners for optimization and constraint verification. Unlike prior work that uses code to interface with robot action modules or pre-designed planners, we steer LLMs to generate code as solvers, planners, and checkers for TAMP tasks requiring symbolic computing, while still leveraging textual reasoning to incorporate common sense. With a multi-round guidance and answer evolution framework, the proposed Code-as-Symbolic-Planner improves success rates by average 24.1\% over best baseline methods across seven typical TAMP tasks and three popular LLMs. Code-as-Symbolic-Planner shows strong effectiveness and generalizability across discrete and continuous environments, 2D/3D simulations and real-world settings, as well as single- and multi-robot tasks with diverse requirements. See our project website\footnote[2]{https://yongchao98.github.io/Code-Symbol-Planner/\label{website}} for prompts, videos, and code.
\end{abstract}

\section{INTRODUCTION} \label{sec:introduction}
Enabling agents to find and execute optimal plans for complex tasks is a long-standing goal in robotics. Robots must reason about the environment, identify a valid sequence of actions, and ensure those actions are feasible given the robot’s motion capabilities—a challenge known as task and motion planning (TAMP) \cite{integrated-TAMP}. Traditional approaches specify tasks using formal representations, such as PDDL \cite{pddl_reference} or temporal logic \cite{temporal_logic_reference}, which are then solved with dedicated planners. While effective, these representations require significant expertise, making them unsuitable for non-expert users. Furthermore, they lack generalizability, as different tasks often require different specialized planners.

Pre-trained Large Language Models (LLMs) have shown remarkable performance on many language-related tasks \cite{llms-few-shot-learners}, driving a surge of interest in applying them to robot TAMP \cite{text2motion, llm-grop}. A common approach \cite{llms-few-shot-learners, errors-are-useful-prompts, llms-zero-shot-reasoners, llms-construct-and-utilize-world-models-for-task-planning} uses LLMs to select actions from pre-defined skill primitives, completing tasks step by step with text or code as intermediates, such as SayCan \cite{saycan}, Inner Monologue \cite{inner-monologue}, Code-as-Policies \cite{code-as-policies}, and ProgPrompt \cite{progprompt}. However, these methods rely solely on the LLM's common-sense reasoning, without leveraging symbolic computation, making them unreliable for complex TAMP tasks involving numeric constraints, optimization, search, and logic. Another line of work integrates LLMs with traditional TAMP planners, such as Text2Motion \cite{text2motion} and AutoTAMP \cite{autotamp}, where LLMs translate natural language instructions into formal representations for downstream TAMP solvers. While these methods incorporate symbolic reasoning, their generalizability is limited, as each task requires carefully designed planners and customized integration frameworks.

To address the limitations of existing LLM-based TAMP approaches, we propose a new paradigm: querying LLMs to directly generate code that serves as both the robot’s TAMP planner and checker. We refer to this approach as Code-as-Symbolic-Planner. Unlike Code-as-Policies \cite{code-as-policies}, which uses LLMs to generate code as intermediates for connecting with action modules—relying primarily on commonsense reasoning without efficient symbolic computation, Code-as-Symbolic-Planner generates code that explicitly performs efficient search and reasoning over valid plans, incorporating symbolic computation into the planning process. Since LLMs can flexibly generate diverse codes for different tasks, this approach naturally integrates symbolic computation into TAMP while preserving broad generalizability.

Our initial tests of querying LLMs to generate symbolic code reveal: (1) Current LLMs demonstrate the ability or at least the potential to generate complete symbolic code capable of solving TAMP tasks. (2) However, LLMs often produce inconsistent code versions, many of which are ineffective or lack proper symbolic computation. To overcome this challenge, we propose a multi-agent, multi-round guidance framework, where the same type of LLM acts as the coding guide, code generator, and answer checker. Through iterative self-guidance, generation, and verification, the symbolic code is progressively refined until a correct solution for the TAMP task is achieved. We also measure code complexity using rule-based methods to provide more effective guidance.

We evaluate Code-as-Symbolic-Planner on seven representative TAMP tasks, including Blocksworld and Path Planning, using three popular LLMs (GPT-4o, Claude3-5-Sonnet, and Mistral-Large). These tasks cover both complex discrete task planning and continuous motion planning in 2D/3D simulations and real environments. Compared to four baseline methods using LLMs as direct planners and a method using OpenAI Code Interpreter for code generation, Code-as-Symbolic-Planner improves success rates by an average of 24.1\% over the best baseline and shows much better scalability on highly complex tasks. This work highlights the potential of a third paradigm for LLM-based TAMP: directly generating symbolic code as the TAMP planner.

\begin{figure}[!htbp]
  \centering
  \includegraphics[width=0.85\linewidth]{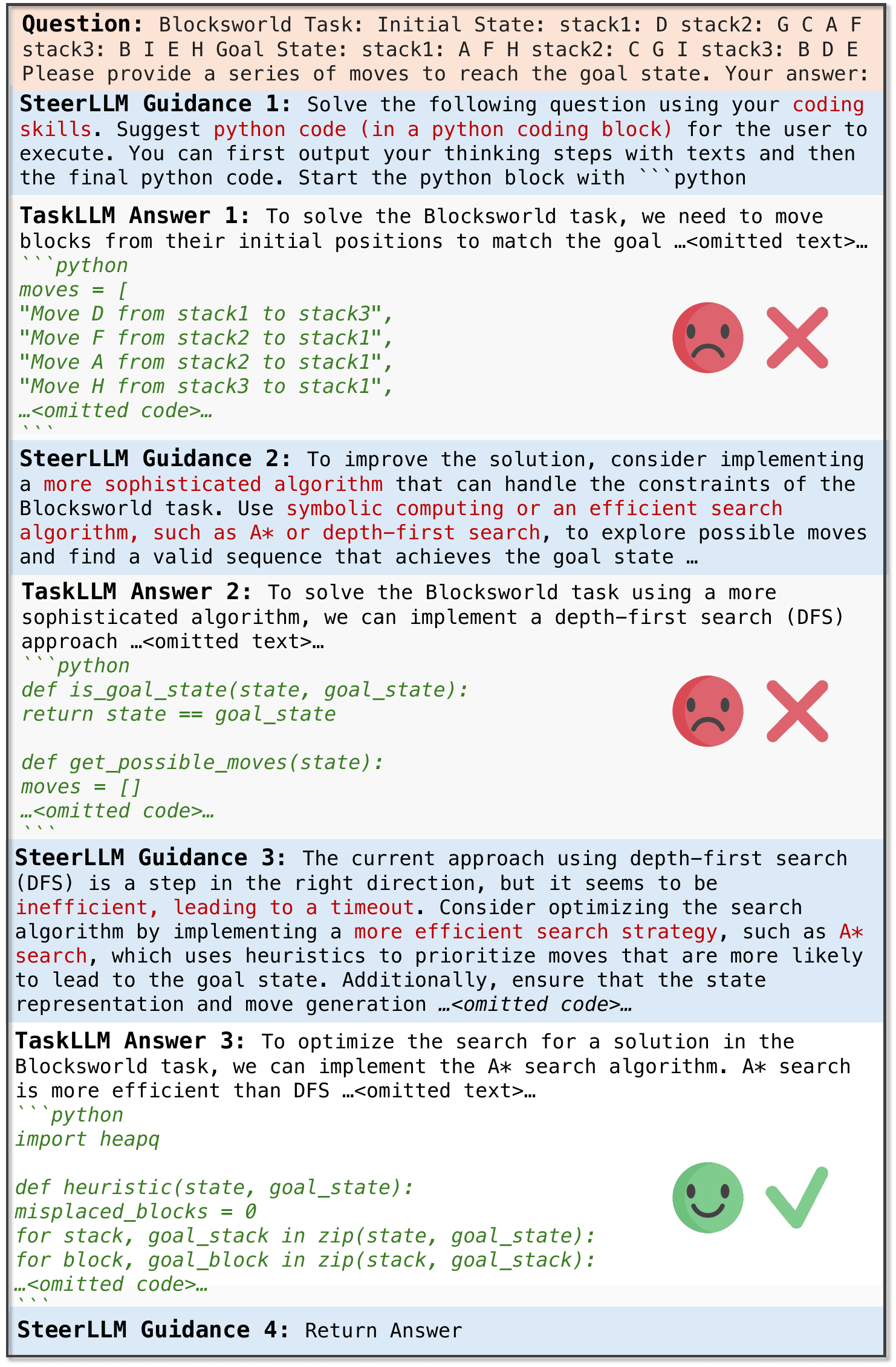}
  \caption{Example of Code-as-Symbolic-Planner applied to symbolic code generation for Blocksworld task planning. In each round, the same LLM checks the current answer, reviews previous answers, self-reflects to generate guidance for the next round, and produces a new code solution. Code-as-Symbolic-Planner returns the final answer once the LLM determines it is complete and correct.}
  \label{fig:Code-as-Symbolic-Planner}
\end{figure}

\section{RELATED WORK} \label{sec:related-work}
\textbf{Task and Motion Planning} Robotic planning spans both high-level discrete task planning \cite{strips-reference} and low-level continuous motion planning \cite{lavalle2006planning}, collectively known as task and motion planning \cite{integrated-TAMP}. Existing approaches typically follow one of three strategies: (1) ensuring motion feasibility before sequencing actions \cite{ferrer2017combined}, (2) generating action sequences first, then resolving motion constraints \cite{lagriffoul2016combining,garrett2020pddlstream}, or (3) interleaving task and motion reasoning \cite{colledanchise2019towards}. However, these methods are inherently limited to goals expressible using predefined predicates and executable through a fixed set of symbolic operators~\cite{sun2022MILP, katayama2020fast}.

\textbf{LLMs for TAMP} The strong reasoning abilities of LLMs \cite{llms-few-shot-learners,llms-zero-shot-reasoners} have sparked interest in using them for TAMP. One line of work directly uses LLMs as task planners \cite{llms-zero-shot-planners, text2motion, tidybot, progprompt, scalable-multi-robot}, requiring additional primitive actions to interface with motion control policies. Other works query LLMs to generate sub-task sequences in code format, which are then executed by downstream motion modules~\cite{code-as-policies,progprompt}. Another approach translates natural language instructions into formal representations or reward functions for existing TAMP solvers~\cite{autotamp,yang-VLM-PDDL} or reinforcement learning systems~\cite{eureka}. However, none of these approaches directly leverage LLMs to generate symbolic code that directly acts as TAMP planner and solver or constraint checker.

\textbf{LLMs for Code Generation} Modern LLMs are extensively trained on diverse code datasets~\cite{gpt-4,llama-3-report}. Recent work has focused on optimizing agent frameworks, improving training strategies, and simplifying tasks to better utilize LLMs as capable software developers~\cite{swebench,opendevin}. Another line of research explores using LLMs to generate code for solving mathematical and logical problems~\cite{mathcoder,steeringlargelanguagemodels}. In this work, we investigate the potential and challenges of using LLM-generated code to solve TAMP tasks.

\section{PROBLEM DESCRIPTION} \label{sec:problem-description}
This work addresses task and motion planning for both single- and multi-robot systems, assuming the LLM has full knowledge of the environment and each robot's capabilities. To provide task goals and observations to the LLM, we manually define functions that translate them into text prompts.

The goal is to convert natural language instructions—including spatial, logical, and temporal constraints—into task plans (sequences of pre-defined actions) or motion plans (sets of timed waypoints, e.g., ($x_i, y_i, t_i$)). For motion planning, the environment is represented as named obstacles described by polygons with locations and orientations. Drones, racecars, and robot arms must stay within maximum velocity limits, and total operation time must not exceed the task’s time limit. Full motion trajectories are assumed to be linear interpolations between waypoints, with complex trajectories achievable through denser waypoint sequences. The planned output of LLMs should be the sequence of waypoints and timepoints for each robot or drone.

\begin{figure}[!htbp]
  \centering
  \includegraphics[width=0.9\linewidth]{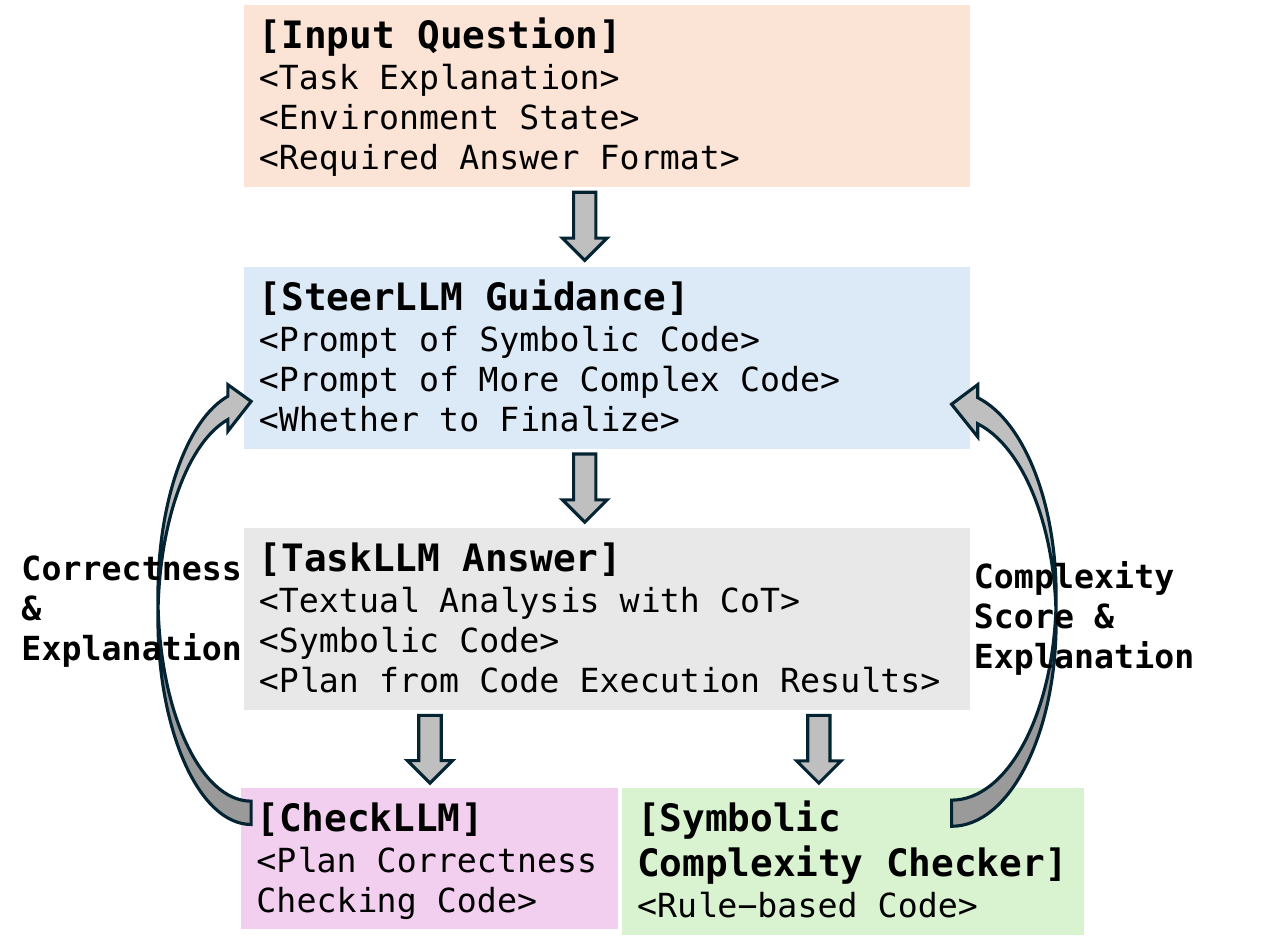}
  \caption{High-level structure of the framework for Code-as-Symbolic-Planner. The arrows on the left and right represent self-reflection on the correctness of the plan and the symbolic complexity of the code, followed by re-guidance and regeneration in the multi-round setting.}
  \label{fig:RoboSteer_framework}
\end{figure}

\section{METHODS} \label{sec:methods}
\begin{figure*}[!htbp]
  \centering
  \includegraphics[width=0.8\linewidth]{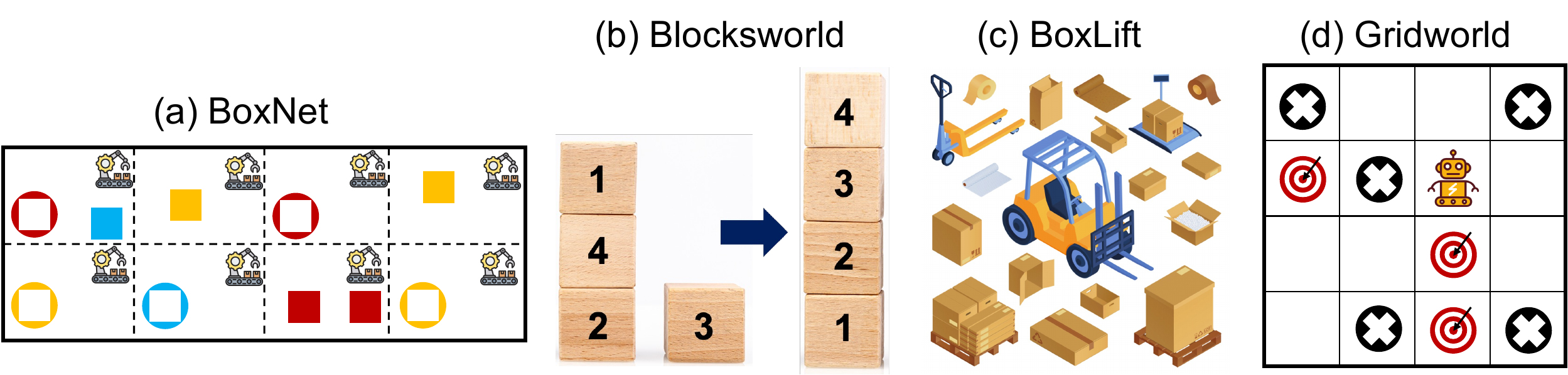}
  \caption{Four robot task planning tasks in discrete states: (a) BoxNet, (b) Blocksworld, (c) BoxLift, (d) Gridworld.}
  \label{fig:2D_simulation}
\end{figure*}

\begin{figure*}[!htbp]
  \centering
  \includegraphics[width=0.8\linewidth]{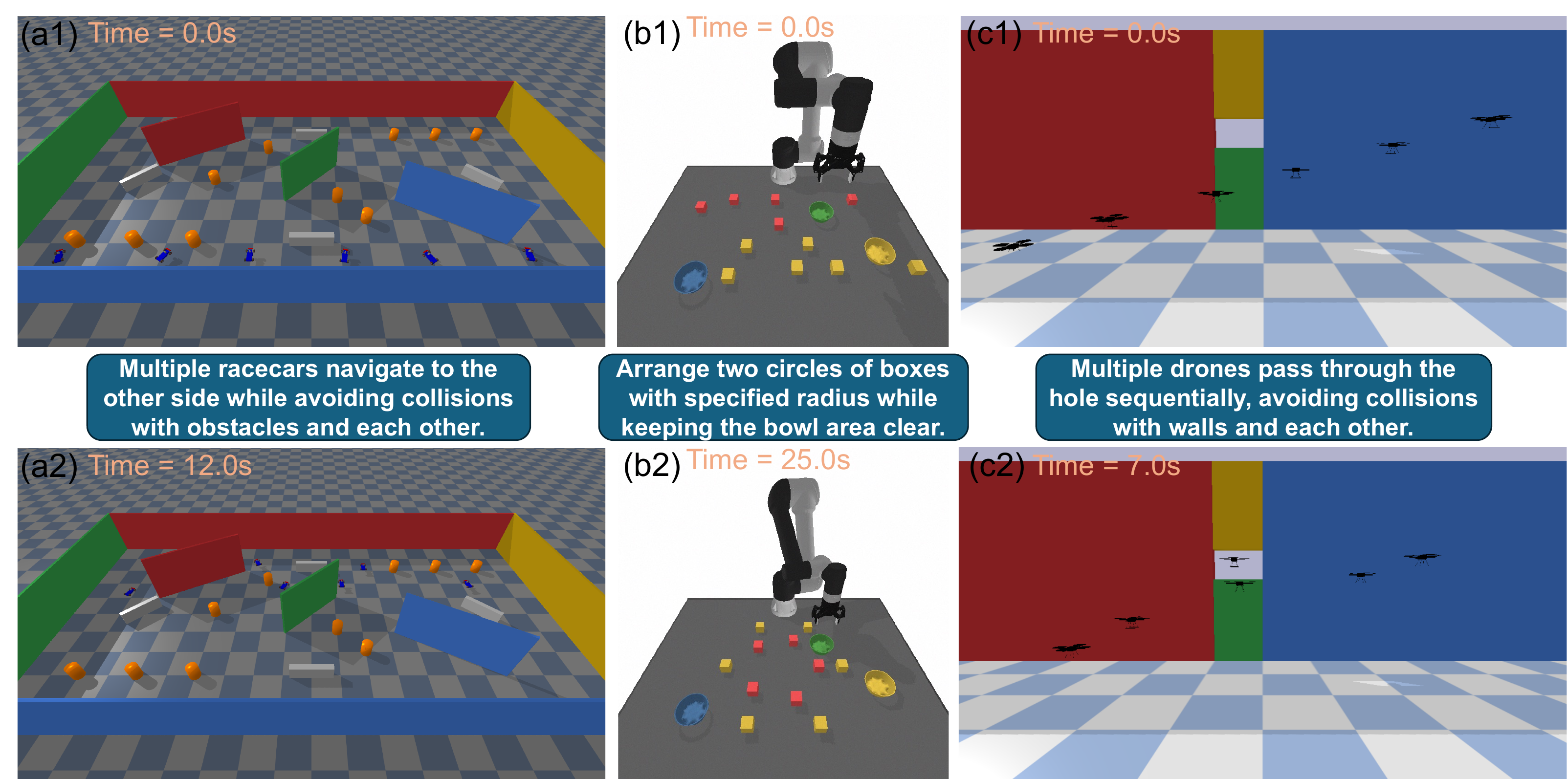}
  \caption{Three robot motion planning tasks in continuous states: (a) Path-Racecars, (b) Shape Formation, (c) Path-Drones.}
  \label{fig:3D_simulation}
\end{figure*}

We compare Code-as-Symbolic-Planner with other five LLM-based TAMP methods, focusing on keeping generalizability across tasks and requirements. Specifically, we evaluate against four baselines with LLMs as the direct planner and one baseline using OpenAI Code Interpreter for symbolic code generation.

\subsection{Code-as-Symbolic-Planner} \label{subsec:Code-as-Symbolic-Planner}
Figure~\ref{fig:Code-as-Symbolic-Planner} displays one example of Code-as-Symbolic-Planner for Blocksworld task and Figure~\ref{fig:RoboSteer_framework} illustrates the general framework of Code-as-Symbolic-Planner.

\textbf{Non-effective Code Versions} As shown in ‘TaskLLM Answer 1’ of Figure~\ref{fig:Code-as-Symbolic-Planner}, when asked to generate code for the task, the LLM often produces trivial code that simply outputs the plan as text, without effective symbolic computation. This limitation, we believe, hampers current and past attempts to use LLMs for TAMP tasks through coding.

To address this, we design a multi-agent, multi-round framework with self-checking modules. We prompt the same LLM to act as three agents: TaskLLM, which generates complete planning code; SteerLLM, which provides guidance prompts to TaskLLM and decides when to finalize the plan; and CheckLLM, which produces checking code to evaluate the correctness of TaskLLM’s output. CheckLLM’s results are fed back to SteerLLM to help generate guidance. Additionally, we introduce a rule-based Symbolic Complexity Checker to analyze the symbolic computation level in the generated code, with its analysis also fed back to SteerLLM. We include all full prompts and scripts in our webpage\footref{website}. A schematic illustration of each component and the overall flowchart is shown in Figure~\ref{fig:RoboSteer_framework}. We set the maximum number of re-guidance and regeneration rounds to 3, after which the final answer is forced to be accepted.

\textbf{SteerLLM} guides TaskLLM in generating code. To enable effective evaluation of the generated code, SteerLLM is equipped with two key components: CheckLLM, which assesses the correctness of the generated plan, and the Symbolic Complexity Checker, which evaluates whether the generated code appropriately leverages symbolic computing.

\textbf{CheckLLM} generates and executes code that verifies the correctness of the current answer. It then returns the evaluation results along with explanations to SteerLLM. Since many planning tasks benefit from code-based verification, this process often offers a more reliable and objective assessment.

\textbf{Symbolic Complexity Checker} is a rule-based script that analyzes the generated code to detect patterns related to iteration, search processes, numeric operations, permutations, and combinations. It then provides a complexity summary and assigns a complexity score. This helps SteerLLM assess whether the generated code demonstrates sufficient sophistication for the given task. Since TaskLLM often produces text-like code that is prone to errors, the Symbolic Complexity Checker’s assessment serves as valuable input for SteerLLM, though it does not dictate its decisions.

\subsection{Baseline Methods} \label{subsec:methods-llm-task-planning}
A common approach is to query LLMs to directly generate a sequence of sub-tasks or motion waypoints from a given language instruction, i.e., LLM works as the direct planner. We evaluate and compare against four methods that each use LLMs as the direct planner:

\textbf{Only Question} Only input the original question.

\textbf{Code Answer} Prompting LLMs to first analyze the question with Chain-of-Thought (COT)~\cite{CoT} and then output the code answer.

\textbf{SayCan}~\cite{saycan} operates by iteratively prompting an LLM to generate the next sub-task in a sequence, conditioned on the previously generated sub-tasks. The next sub-task is selected from the top
K candidates by combining two likelihoods: (1) the LLM-generated likelihood and (2) the feasibility likelihood of each candidate action. Following the approach proposed in SayCan\cite{saycan}, the sub-task with the highest combined likelihood is chosen. We set K to 5.

\textbf{HMAS-2}~\cite{scalable-multi-robot} is a multi-agent framework designed for multi-robot settings, where each robot is equipped with an LLM to provide local feedback to a central LLM planner. This framework follows an iterative planning approach, generating one sub-task per step. No local feedback is used when applied to single-robot tasks.

Apart from above four baselines, we also compare with \textbf{OpenAI Code Interpreter}\footnote[4]{https://platform.openai.com/docs/assistants/tools/code-interpreter/\label{Code Interpreter}}, which is also designed to generate code to solve tasks when needed, namely integrating symbolic computing for TAMP tasks.

\begin{table*}[!htbp]
\vspace{2mm}
\caption{Evaluation results on GPT-4o. Code-Symbo.-P. refers to Code-as-Symbolic-Planner.}
\centering
\begin{tabular}{|l|cccccc|cc|}
\hline
 Success & \begin{tabular}[c]{@{}c@{}}Code-\\ Symbo.-P.\end{tabular} & \begin{tabular}[c]{@{}c@{}}Only\\ Question\end{tabular} & Code Answer & \begin{tabular}[c]{@{}c@{}}Code\\ Interpreter\end{tabular}
    & SayCan & HMAS-2
    & \begin{tabular}[c]{@{}c@{}}wo Symbolic\\Complexity Checker\end{tabular} & wo CheckLLM \\ \hline
\multicolumn{1}{|l|}{BoxNet} & 74.2\% & 27.5\% & 68.3\% & \textbf{76.7\%} & 42.9\% & 50.7\% & 70.0\% & 71.4\% \\
\multicolumn{1}{|l|}{Blocksworld} & \textbf{48.2\%} & 0.7\%  & 0.4\% & 0.0\% & 0.7\% & 1.4\% & 41.4\% & 39.3\% \\
\multicolumn{1}{|l|}{BoxLift} & 17.6\% & 6.4\% & 9.3\% & \textbf{19.6\%} & 6.8\% & 8.2\% & 14.3\% & 13.2\% \\ 
\multicolumn{1}{|l|}{Gridworld} &\textbf{63.6\%} & 20.0\% & 46.7\% & 30.8\% & 17.9\% & 26.4\% & 55.0\% & 58.6\% \\ 
\multicolumn{1}{|l|}{Path-Racecars} & \textbf{59.3\%} & 18.6\% & 37.9\% & 30.0\% & 17.1\% & 20.0\% & 53.6\% & 52.1\% \\ 
\multicolumn{1}{|l|}{Shape Formation} & \textbf{41.4\%} & 8.6\% & 11.4\% & 37.1\% & 7.9\% & 12.9\% & 7.1\% & 10.0\% \\ 
\multicolumn{1}{|l|}{Path-Drones} & \textbf{62.1\%} & 15.7\% & 39.3\% & 45.7\% & 17.1\% & 20.7\% & 47.9\% & 54.3\% \\ \hline
\multicolumn{1}{|l|}{\textbf{Average}} & \textcolor{cyan}{\textbf{52.3\%}} & 13.9\% & 30.5\% & 34.3\% & 15.8\% & 20.0\% & 41.3\% & 42.7\% \\ \hline
\end{tabular}
\label{tab:gpt-4o-result}
\end{table*}

\begin{table*}[!htbp]
\vspace{2mm}
\caption{Evaluation results on Claude-3-5-sonnet-20241022 (Claude3-5-sonnet) and Mistral-large-latest (Mistral-Large).}
\centering
\begin{tabular}{|l|ccc|ccc|}
\hline
 Success & \begin{tabular}[c]{@{}c@{}}Code-Symbo.-P.\\(Claude3-5-sonnet)\end{tabular} & \begin{tabular}[c]{@{}c@{}}Only Question\\(Claude3-5-sonnet)\end{tabular} & \begin{tabular}[c]{@{}c@{}}Code Answer\\ (Claude3-5-sonnet)\end{tabular} & \begin{tabular}[c]{@{}c@{}}Code-Symbo.-P.\\(Mistral-Large)\end{tabular} & \begin{tabular}[c]{@{}c@{}}Only Question\\(Mistral-Large)\end{tabular} & \begin{tabular}[c]{@{}c@{}}Code Answer\\ (Mistral-Large)\end{tabular} \\ \hline
\multicolumn{1}{|l|}{BoxNet} & 52.2\% & \textbf{67.1\%} & 32.9\% & \textbf{82.1\%} & 76.4\% & 73.6\% \\
\multicolumn{1}{|l|}{Blocksworld} & \textbf{41.8\%} & 0.4\% & 0.7\% & \textbf{27.5\%} & 0.0\% & 0.0\% \\
\multicolumn{1}{|l|}{BoxLift} & \textbf{86.4\%} & 7.1\% & 75.7\% & \textbf{5.7\%} & 2.9\% & 1.8\% \\ 
\multicolumn{1}{|l|}{Gridworld} & \textbf{79.3\%} & 35.7\% & 17.1\% & \textbf{63.6\%} & 7.9\% & 40.7\% \\ 
\multicolumn{1}{|l|}{Path-Racecars} & \textbf{73.6\%} & 40.0\% & 16.4\% & \textbf{51.4\%} & 9.3\% & 37.1\% \\ 
\multicolumn{1}{|l|}{Shape Formation} & \textbf{50.7\%} & 10.7\% & 15.0\% & \textbf{35.7\%} & 5.0\% & 6.4\% \\ 
\multicolumn{1}{|l|}{Path-Drones} & \textbf{63.6\%} & 25.0\% & 43.6\% & \textbf{55.7\%} & 7.9\% & 28.6\% \\ \hline
\multicolumn{1}{|l|}{\textbf{Average}} & \textcolor{cyan}{\textbf{63.9\%}} & 26.6\% & 28.8\% & \textcolor{cyan}{\textbf{46.0\%}} & 15.6\% & 26.9\% \\ \hline
\end{tabular}
\label{tab:other LLM results}
\end{table*}

\begin{figure*}[!htbp]
  \centering
  \includegraphics[width=0.8\linewidth]{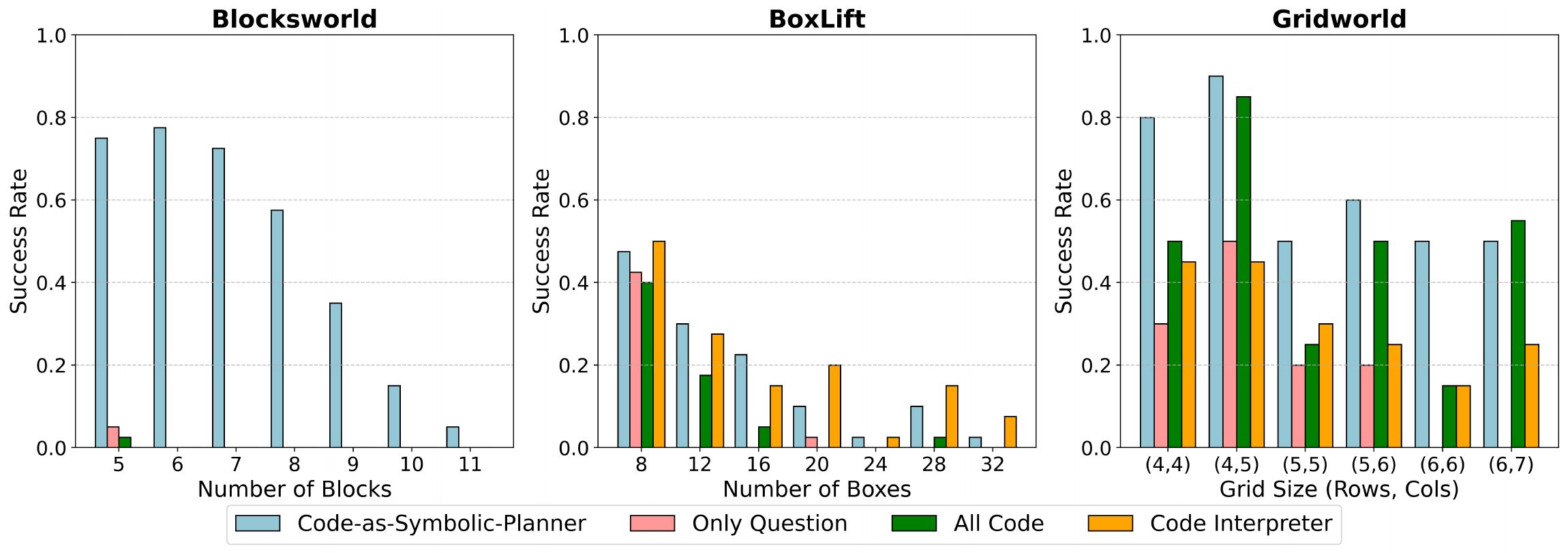}
  \caption{Task success rates across increasing task complexity for the four studied methods on three tasks.}
  \label{fig:varied-complex}
\end{figure*}

\section{EXPERIMENTAL DESIGN} \label{sec:experimental-design}
We compare the LLM-based planning methods on four discrete task planning tasks and three continuous motion planning tasks including both single- and multi-robot settings, as shown in Figure~\ref{fig:2D_simulation} and \ref{fig:3D_simulation}, respectively. In task planning scenarios, such as BoxNet, Blocksworld, BoxLift, and Gridworld, LLMs are required to generate sequences of high-level actions. In motion planning scenarios, such as Path-Racecars, Shape Formation, and Path-Drones, LLMs are required to generate sequences of motion waypoints. In addition to the 2D scenarios for our experiments, we also perform experiments in a 3D environment simulated using Pybullet \cite{pybullet} and real hardware settings with two robot arms, as shown in Figure~\ref{fig:3D_simulation} and \ref{fig:real_experiment}. These tested tasks originate from previous works~\cite{planbench,scalable-multi-robot, nishan, autotamp, promst, gensim}.

Each task includes 140 test samples with diverse complexities and constraints. To evaluate the method's ability to handle highly complex tasks, the environment is intentionally made more challenging than in previous studies. We evaluate the success of each testing trial by checking three criteria: (1) syntax correctness, (2) task completion, and (3) satisfaction of required constraints, such as collision avoidance, time limits, and logical consistency. For each method, the LLM is prompted with a task description without any few-shot examples. If the generated code takes more than 50 seconds to execute, we terminate the trial and record it as a failure. The specific descriptions of the testing tasks are as follows:

\textbf{BoxNet~\cite{scalable-multi-robot}} In Figure~\ref{fig:2D_simulation}(a), the environment consists of cell regions, robot arms, colored boxes, and corresponding colored goal locations. The objective is to move each box to its matching goal location in the fewest time steps. Each robot arm is restricted to the cell it occupies and can perform two actions: (1) move a box to a neighboring cell or (2) place a box in a goal location within the same cell.

\textbf{Blocksworld~\cite{llm+p}} In Figure~\ref{fig:2D_simulation}(b), the goal is to stack blocks in a specified order. The robot can take four actions: (1) pick up a block, (2) unstack a block from another block, (3) put down a block, and (4) stack a block on another block.

\textbf{BoxLift~\cite{scalable-multi-robot}} In Figure~\ref{fig:2D_simulation}(c), robots are tasked with lifting boxes in the fewest time steps. Each robot has a different lifting capacity, and each box has a different weight. Multiple robots can collaborate to lift the same box in a single time step. A box is successfully lifted when the combined lifting capacity of the assigned robots exceeds the box's weight.

\textbf{Gridworld~\cite{promst}} Figure~\ref{fig:2D_simulation}(d) consists of obstacles (black) and goals (red). The robot needs to visit all goals. Attempts to move into obstacles or move out of the grid will result in failure. The robot has five possible actions: (1) move up, (2) move down, (3) move left, (4) move right, (5) visit goal.

\textbf{Path-Racecars~\cite{autotamp}} Figure~\ref{fig:3D_simulation}(a) queries LLMs to plan the racecar trajectory waypoints to move all the racecars from one side to another under varied environments and obstacles.

\textbf{Shape Formation~\cite{gensim}} Figure~\ref{fig:3D_simulation}(b) queries LLMs to plan the picking order and placing positions of randomly located colored boxes to form required shapes like circles and triangles. The locations and shapes of bowls are unchangeable so that the boxes should not occupy bowls' areas.

\textbf{Path-Drones~\cite{autotamp}} Figure~\ref{fig:3D_simulation}(c) requires LLMs to plan the drone trajectory waypoints to pass through the hole one by one while keeping safe distances to the wall and other drones.

\section{RESULTS} \label{sec:results}
Table~\ref{tab:gpt-4o-result} presents the task success rates for all methods across seven tasks using GPT-4o. Code-as-Symbolic-Planner outperforms the four baselines where LLMs act as direct planners. Compared to the OpenAI Code Interpreter, which also incorporates symbolic computing, Code-as-Symbolic-Planner achieves an average success rate improvement of 18\%. Additionally, we evaluate Claude3.5-sonnet and Mistral-Large, with results shown in Table~\ref{tab:other LLM results}. These results further demonstrate the notably better performance of Code-as-Symbolic-Planner compared to the tested baselines.

\textbf{Scalability to Higher Complexity} Figure~\ref{fig:varied-complex} shows the success rates of each method as task complexity increases, with more objects and larger playgrounds. Among the four methods, the Only Question approach, where LLMs serve as direct planners, experiences the fastest performance decline, highlighting the benefits of integrating symbolic computing.

Notably, Code-as-Symbolic-Planner shows the best scalability on more complex tasks compared to the other three baselines. Ideally, the symbolic code it generates should handle these tasks effectively. 

\textbf{Failure Reasons of Code-as-Symbolic-Planner} However, in some highly complex cases, the generated code exceeds the 50-second execution time limit, contributing to its performance drop in the most challenging scenarios. This also demonstrates one typical failure case of Code-as-Symbolic-Planner that sometimes the generated code does not utilize the most efficient algorithms or the task is too complex so that purely algorithm-based methods are not sufficient. Another major bottleneck we identified is the LLM's inability to translate real-world application problems into abstract optimization or logistical tasks solvable through code, which is also challenging to human experts.

\begin{figure}[!htbp]
  \vspace{1mm}
  \centering
  \includegraphics[width=0.9\linewidth]{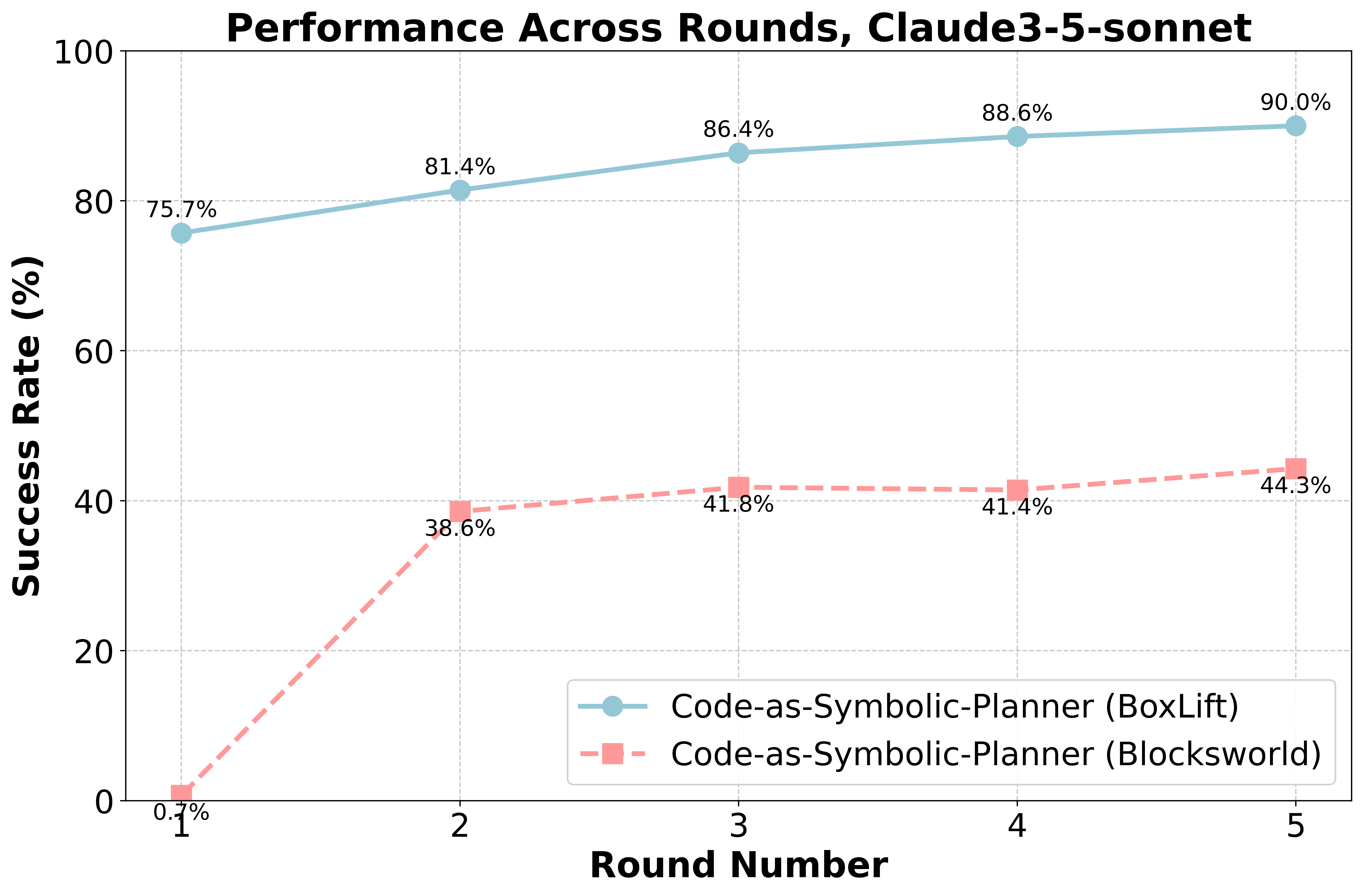}
  \caption{Success rates of Code-as-Symbolic-Planner with Claude3-5-sonnet across different maximum numbers of guidance and generation rounds. The blue and red lines represent tasks BoxLift and Blocksworld, respectively.}
  \label{fig:vary_rounds}
\end{figure}

\textbf{Impacts of Maximum Guidance/Generation Rounds} In our study, we set the maximum iteration rounds for Code-as-Symbolic-Planner to 3. To examine the impact of this setting, Figure~\ref{fig:vary_rounds} shows the success rates for BoxLift and Blocksworld tasks as the maximum rounds increase from 1 to 5. Note that with only 1 round, Code-as-Symbolic-Planner functions identically to the Code Answer method. 

We observe that success rates generally improve as the number of rounds increases, but they gradually plateau once the round number exceeds 3. This result highlights the potential for Code-as-Symbolic-Planner to achieve better performance with additional iterations, while also supporting the reasonableness of our decision to set the maximum round number to 3 in this study. Meanwhile, it proves the multi-round setting in Code-as-Symbolic-Planner is reasonable.

\textbf{Effects of CheckLLM and Symbolic Complexity Checker} In Table~\ref{tab:gpt-4o-result}, we evaluate the impact of two key components CheckLLM and Symbolic Complexity Checker by measuring the performance of Code-as-Symbolic-Planner when each component is removed. We observe that the average success rate drops by 9.6\% without CheckLLM and by 11.0\% without the Symbolic Complexity Checker. These results demonstrate the effectiveness of both components in enhancing the guidance provided by SteerLLM.

\begin{figure}[!htbp]
  \vspace{1mm}
  \centering
  \includegraphics[width=0.85\linewidth]{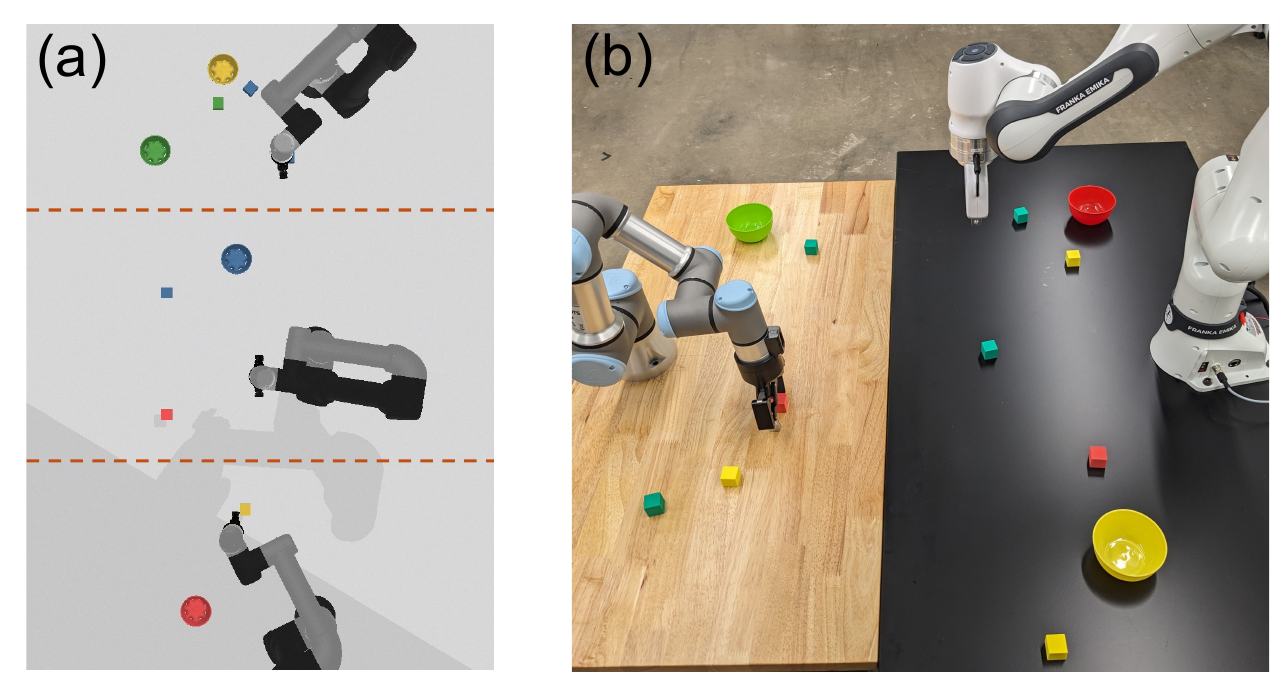}
  \caption{3D simulation environments and real hardware demonstrations: robot arms collaborate to move all the boxes into the same colored bowls. Each robot arm has a limited workspace and can only move within its assigned region (divided by the dotted lines or different colored tables).}
  \label{fig:real_experiment}
\end{figure}

\begin{table}[!htbp]
\vspace{2mm}
\caption{Evaluation results on BoxNet for real robot arms and 3D simulations, reported as average success rates over 20 runs.}
\label{tab:real_env_table}
\resizebox{\columnwidth}{!}{
\begin{tabular}{|l|c|c|c|}
\hline
    Success & Code-Symbo.-P. & Only Question & Code Answer \\ \hline
    Two Robots (real) & \textbf{100\%} & 45\% & 70\% \\
    Three Robots (3D) & \textbf{100\%} & 20\% & 55\% \\
    Six Robots (3D) & \textbf{95\%} & 0\% & 35\% \\ \hline
\end{tabular}
}
\end{table}

\textbf{3D Simulation and Physical Demonstrations} For the BoxNet task, we conduct experiments in both a 3D simulation environment using Pybullet~\cite{pybullet} and on a real two-arm robotic system, as shown in Figure~\ref{fig:real_experiment}. The environment includes colored boxes, colored bowls, and robot arms. The objective is to move each colored box into the bowl of the matching color using the fewest possible actions. Each robotic arm is stationary and limited to operating within its designated workspace (indicated by the dotted lines or the different colored tables). Arms can only pick up and place boxes within their workspace or on the border between workspaces. We also evaluate the system’s scalability by testing with two arms in the real-world experiments, and with either three or six arms in the simulation environment.

The 3D real-world and simulation environments introduce additional complexity by incorporating an image-to-text model, ViLD~\cite{ViLD}, which provides object bounding boxes and text descriptions. Additionally, the 3D simulation includes a richer environment model that allows for action execution errors caused by dynamic factors (e.g., boxes slipping from the gripper), requiring real-time re-planning. When such execution errors occur, we query the three tested methods to re-plan the next steps.

We conduct 20 trials for each scenario. The results, shown in Table~\ref{tab:real_env_table}, demonstrate that Code-as-Symbolic-Planner consistently outperforms methods that do not incorporate symbolic computing, aligning with the findings from the other experiments.

\section{CONCLUSION} \label{sec:conclusion}
We present Code-as-Symbolic-Planner, a framework that leverages LLMs to generate symbolic code for general robot TAMP. Compared to the two existing types of approaches for applying LLMs to TAMP—using LLMs either as direct planners or as translators interfacing with external planners, Code-as-Symbolic-Planner directly integrates symbolic computing into the planning process while maintaining strong generalizability across diverse tasks and constraints.

Our experiments, conducted on seven representative TAMP tasks in both 2D/3D simulation environments and real hardware settings, demonstrate the strong potential of Code-as-Symbolic-Planner in solving complex robot TAMP problems. We believe this framework offers a promising third paradigm for applying LLMs to robot TAMP tasks.

\textbf{Limitations and Future Directions}
First, while our results show that current LLMs can generate correct code to solve TAMP tasks under multi-round guidance, the generated code is often not optimal, especially for high-complexity tasks where it may exceed time limits. Generating planners with coding also risks adding errors. In robotics, many existing planners are highly efficient but lack generalizability. Combining Code-as-Symbolic-Planner with these efficient planners is a promising direction for future work.

Second, helping LLMs formalize tasks into code remains a key challenge. In our study, all task information and environment states are provided as detailed text descriptions. To improve usability, future work could explore incorporating additional input modalities, such as vision and force sensing.

Third, We also observe that the required output format for plans strongly influences the coding success rate, underscoring the need to enhance LLMs' self-formalization capabilities for TAMP problems.

\clearpage

\addtolength{\textheight}{-12cm}   





\bibliographystyle{IEEEtran.bst}
\bibliography{references}

\end{document}